\documentclass[10pt,twocolumn,letterpaper]{article}

\usepackage{cvpr}
\usepackage{times}
\usepackage{epsfig}
\usepackage{graphicx}
\usepackage{amsmath}
\usepackage{amssymb}
\usepackage{mathtools}
\usepackage{multirow}
\usepackage{enumitem}
\usepackage{color}

\newcommand{\keypoint}[1]{\vspace{0.1cm}\noindent\textbf{#1}\quad}
\newcommand{\cut}[1]{}
\usepackage{algorithm}
\usepackage{algpseudocode}

\usepackage[nopar]{lipsum}
\usepackage{arydshln}
\DeclareMathAlphabet\mathbfcal{OMS}{cmsy}{b}{n}
\newcommand{\modelName}[1]{Sketch2Vec}

\makeatother
\usepackage[pagebackref=true,breaklinks=true,letterpaper=true,colorlinks,bookmarks=false]{hyperref}

\cvprfinalcopy 

\def\cvprPaperID{2542} 

\begin{document}

\title{Vectorization and Rasterization: Self-Supervised Learning \\
for Sketch and Handwriting}
\author{Ayan Kumar Bhunia\textsuperscript{1} \hspace{.2cm} Pinaki Nath Chowdhury\textsuperscript{1,3}  \hspace{.2cm} Yongxin Yang\textsuperscript{1,3} \hspace{.2cm} Timothy M. Hospedales\textsuperscript{1,2}
\\ Tao Xiang\textsuperscript{1,3}\hspace{.2cm}  Yi-Zhe Song\textsuperscript{1,3} \\
\textsuperscript{1}SketchX, CVSSP, University of Surrey, United Kingdom \hspace{.08cm} \textsuperscript{2}University of Edinburgh, United Kingdom.\\
\textsuperscript{3}iFlyTek-Surrey Joint Research Centre on Artificial Intelligence.\\
{\tt\small \{a.bhunia, p.chowdhury, yongxin.yang, t.xiang, y.song\}@surrey.ac.uk, t.hospedales@ed.ac.uk}
}

\maketitle
\ifcvprfinal\thispagestyle{empty}\fi

\begin{abstract}
Self-supervised learning has gained prominence due to its efficacy at learning powerful representations from unlabelled data that achieve excellent performance on many challenging downstream tasks. However, supervision-free pre-text tasks are challenging to design and usually modality specific. Although there is a rich literature of self-supervised methods for either spatial (such as images) or temporal data (sound or text) modalities, a common pre-text task that benefits both modalities is largely missing. In this paper, we are interested in defining a self-supervised pre-text task for sketches and handwriting data. This data is uniquely characterised by its existence in dual modalities of  rasterized images and vector coordinate sequences. We address and exploit this dual representation by proposing two novel cross-modal translation pre-text tasks for self-supervised feature learning: Vectorization and Rasterization. Vectorization learns to map image space to vector coordinates and  rasterization maps vector coordinates to image space. We show that our learned encoder modules benefit both raster-based and vector-based downstream approaches to analysing hand-drawn data. Empirical evidence shows that our novel pre-text tasks surpass existing single and multi-modal self-supervision methods.
\end{abstract}

\vspace{-0.6cm}
\section{Introduction}
\vspace{-0.1cm}

Deep learning architectures \cite{yu2016sketchAnet, he2016deep} have become the de-facto choice for most computer vision applications. However, their success heavily depends on access to large scale labelled datasets \cite{russakovsky2015imagenet} that are both costly and time-consuming to collect. In order to alleviate the data annotation bottleneck, many unsupervised methods \cite{noroozi2016unsupervised, kingma2013auto, doersch2015unsupervised, caron2018deep, he2020momentum} propose to pre-train a good feature representation from large scale unlabelled data. A common approach is to define a \emph{pre-text task} whose labels can be obtained free-of-cost, e.g. colorization \cite{zhang2016colorful}, jigsaw solving \cite{noroozi2016unsupervised}, image rotation prediction \cite{gidaris2018unsupervised}, etc. The motivation is that a network trained to solve such a pre-text task should encode high-level semantic understanding of the data that can be used to solve other downstream tasks like classification, retrieval, etc. Apart from traditional object classification, detection or semantic segmentation,  self-supervision has been extended to sub-domains like human pose-estimation \cite{kocabas2019self}, co-part segmentation \cite{hung2019scops}, and depth estimation \cite{godard2019digging}.

\begin{figure}[]
\begin{center}
  \includegraphics[width=0.9\linewidth]{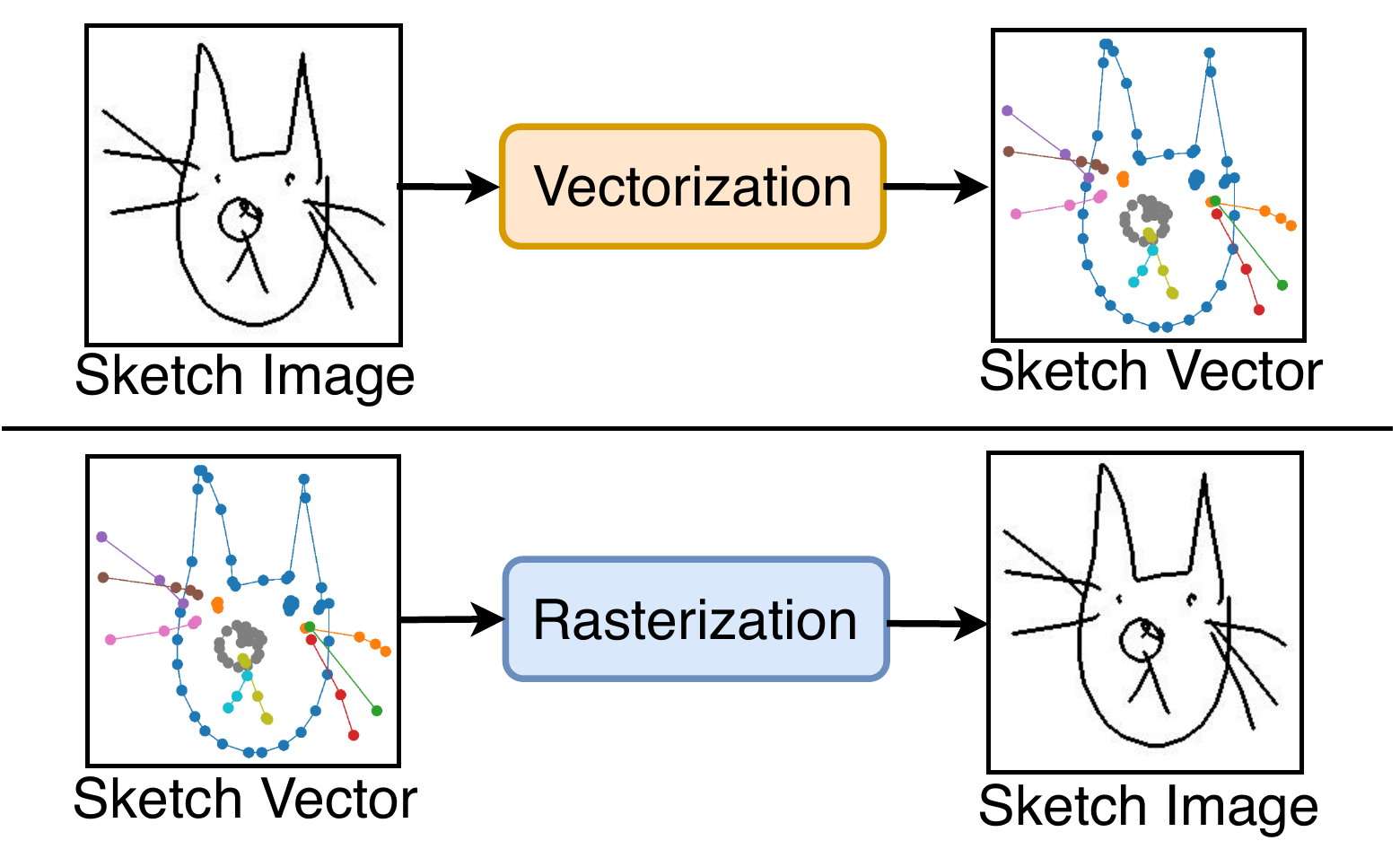}
\end{center}
\vspace{-0.5cm}
  \caption{Schematic of our proposed self-supervised method for sketches. Vectorization drives representation learning for sketch images; rasterization is the pre-text task for sketch vectors.}
  \vspace{-0.6cm}
\label{fig:Self_fig1.pdf}
\end{figure}

In this paper, we propose a self-supervised method for a class of visual data that is distinctively different than photos: sketches \cite{yu2016sketchAnet, sangkloy2016sketchy} and handwriting \cite {poznanski2016cnn} images.  Although sketch and handwriting have been studied as two separate topics by different communities, there exists an underlying similarity in how they are captured and represented. More specifically, they are both recorded as the user's pen tip follows a trajectory on the canvas, and rendered as sparse black and white lines in image space. Both are abstract, in the sense that the same object or grapheme can be drawn in many possible ways \cite{song2018learning, ha2017neural}; while sketch in particular poses the challenge of variable levels of detail~\cite{BMVC_hierarchy} depicted.
Both sketch \cite{xu2018sketchmate} and handwriting \cite{graves2008novel} can be represented in rasterized pixel space, or as a temporal point sequence \cite{ha2017neural}. While each modality has its own benefits, we propose a novel self-supervised task that takes advantage of this dual image/vector space representation. In particular, we use cross modal translation between image and vector space as a self-supervised task to improve downstream performance using either representation (Figure~\ref{fig:Self_fig1.pdf}).

Most existing self-supervised methods are defined for single data modalities. Existing methods for images \cite{chen2020simple, he2020momentum, gidaris2018unsupervised} or videos \cite{korbar2018cooperative} are designed for pixel perfect renderings of scenes or objects, and as such are not suited for sparse black and white handwritten images. For example colorization \cite{zhang2016colorful} and super-resolution \cite{ledig2017photo} pre-texts, and augmentation strategies such as color distortion, brightness, and hue adjustment used by state of the art contrastive methods \cite{he2020momentum, chen2020simple, grill2020bootstrap} --  are not directly applicable to line drawings. For vector sequences, self-supervised methods typically addressed at speech such as Contrastive Predictive Coding (CPC) \cite{henaff2019data} could be used off-the-shelf but do not explicitly handle the stroke-by-stroke nature of handwriting. \cut{On the other hand}Conversely, BERT-like pre-training strategies have \cut{been applied with}had some success with vector-modality sketches \cite{sketchbert2020} but cannot be applied to image-modality sketches. In contrast, our framework can be used to learn a powerful representation for both image and vector domain sketch analysis tasks.
Although a multi-view extension of contrastive learning for self-supervision~\cite{tian2019contrastiveCMC} has been attempted,  we show empirically that our cross-view rasterization/ vectorization synthesis approach provides a superior self-supervision strategy.

In summary, we design a novel self-supervised framework that exploits the dual raster/vector sequence nature of sketches and handwritten data through cross-modal translation (Figure~\ref{fig:Self_fig1.pdf}). Our cross-modal framework is simple and easy to implement from off-the-shelf components. Nevertheless it learns powerful representations for both raster and vector represented downstream sketch and handwriting analysis tasks. Empirically, our framework surpasses state of the art self-supervised methods and approaches and sometimes surpasses the fully supervised alternatives.

\vspace{-0.1cm}
\section{Related Works}
\vspace{-0.1cm}
\keypoint{Sketch Representation Learning:} Learning good sketch representations benefits a variety of sketch-specific problems like classification \cite{yu2016sketch}, retrieval \cite{xu2018sketchmate}, scene understanding \cite{liu2020scenesketcher}, sketch based image retrieval \cite{bhunia2020sketch, pang2019generalising, pang2020solving, dey2019doodle, dutta2019semantically, semi-fgsbir, stylemeup, BMVC_hierarchy}, generative sketch modelling \cite{song2018learning, ha2017neural, sketchxpixelor} etc. While photos are pixel perfect depictions represented by 2D spatial matrices, sketches can be described either as 2D static pixel level \emph{rasterized images} or vector sketches with an ordered sequence of \emph{point coordinates}. Typically, sketch images are processed by convolutional neural networks \cite{bhunia2020sketch, dey2019doodle}, whereas vector sketches needs Recurrent Neural Networks (RNNs) or Transformers \cite{xu2018sketchmate, sampaio2020sketchformer, sketchbert2020} for sequential modelling. There exists no consensus on which sketch modality (image or vector) is better than the other, as each has its own merits  based on the application scenario. While rasterized sketch images are usually claimed to be better for driving  fine-grained retrieval \cite{yu2016sketch, song2017deep}, they fail to model the varying level of abstraction \cite{song2018learning} in the sketch generative process. Conversely, vector sketches are more effective to simulate the human sketching style \cite{ha2017neural} for generation, however, it fails \cite{bhunia2020sketch} for fine-grained instance level image retrieval. From a  computational cost perspective, coordinate based models provides faster cost-effective real-time performance \cite{xu2019multi} for sketch-based human-computer interaction, compared to using rasterized sketch images that impose a costly rendering step and transfer cost of a large pixel array. Attempts have been directed towards combining representations from both sketch images and vector sketches for improved performance in sketch hashing and category level sketch-based image retrieval \cite{xu2018sketchmate}. 

Image-Net pretrained weights are widely used to initialise standard convolutional networks for sketch images, with the first self-supervised alternatives specifically designed for raster sketch images being proposed recently \cite{pang2020solving}. Vector sketches relying on RNN or Transformer do not have the ImageNet initialization option. Thus, Lin \etal \cite{sketchbert2020} employ BERT-like self-supervised learning on vector sketches. 
Nevertheless, these existing self-supervised are proposed for specific modalities (raster image vs vector sketches) and do not generalize to each other. We therefore propose a unified pipeline that leverages this \emph{dual representation of sketches} to learn powerful features for encoding sketches represented in both vector and raster views.

\keypoint{Handwriting Recognition:} Similar to sketches, handwriting recognition has also been heavily explored involving both image space (`offline' recognition) \cite{poznanski2016cnn, bhunia2019handwriting, bluche2016joint, luo2020learn, wang2020decoupled} and vector space (`online' recognition) \cite{hu1996hmm, graves2008novel}. Unlike sketch, there is a consensus in the handwriting community \cite{carbune2020fast, keysers2016multi} that vector representation of handwriting provides better recognition accuracy over offline images. Connectionist temporal Classification (CTC) criterion by Graves \etal \cite{graves2008novel} made end-to-end sequence discriminative learning possible. Following this seminal work \cite{graves2008novel}, earlier handcrafted feature extraction methods \cite{bengio1995lerec, hu1996hmm, almazan2014word} in both the modalities have now been replaced by data driven feature learning  \cite{bhunia2019handwriting, poznanski2016cnn, carbune2020fast} as in many computer vision domains. Nevertheless, data scarcity still remains a bottleneck for both offline and online handwriting recognition, despite advances such as modelling handwriting style variation via adversarial feature deformation module \cite{bhunia2019handwriting} or learning an optimal augmentation strategy \cite{luo2020learn} using reinforcement learning. We demonstrate the first use of self-supervision to improve both offline and online handwriting recognition.

\keypoint{Self-supervised Learning:} Self-supervised learning is now a large field, too big to review in detail here, with recent surveys \cite{jing2020selfSup} providing a broader overview.
\cut{Most self-supervised methods for representation learning can be broadly classified into two groups: generative and discriminative. Generative approaches aim to model the distribution over data and the learned latent embedding is used as image representation. Existing generative approaches includes Auto-Encoders \cite{vincent2008extracting}, Variational Auto-Encoders \cite{kingma2013auto}, Generative Adversarial Networks \cite{goodfellow2014generative} to name a few, which can learn meaningful latent representation along with the generative process.}
As a brief review: Generative models such as VAEs \cite{kingma2013auto} learn representations by modelling the distribution of the data. Contrastive learning \cite{grill2020bootstrap, chen2020simple, he2020momentum} aims to learn discriminative features by minimising the distance between different augmented views of the same image while maximizing it for views from different images. Clustering based approaches \cite{caron2018deep} first cluster the data based on the features extracted from a network, followed by re-training the same network using the cluster-index as pseudo-labels for classification.

Different pre-text tasks have been explored for self-supervised feature learning in imagery, e.g., image colorization \cite{zhang2016colorful},  super-resolution \cite{ledig2017photo}, solving jigsaw puzzles \cite{noroozi2016unsupervised,pang2020solving},  in-painting \cite{pathak2016context}, relative patch location prediction \cite{doersch2015unsupervised}, frame order recognition \cite{misra2016shuffle}, etc. Compared to these approaches, our work is more similar to the few approaches addressing multi-modal data. For instance,  pre-text tasks like visual-audio correspondence  \cite{arandjelovic2017look,korbar2018cooperative}, or RGB-flow-depth correspondence \cite{tian2019contrastiveCMC} within vision. However, these approaches use contrastive losses, which raise a host of complex design issues in batch size, batch sampling strategies, and positive/negative balancing \cite{chen2020simple,patacchiola2020selfsupervised, he2020momentum, henaff2019data, tian2019contrastiveCMC} that are necessary to tune, in order to obtain good performance. Furthermore they tend to be extremely expensive to scale due to the ultimately quadratic cost of comparing sample \emph{pairs} \cite{chen2020simple,patacchiola2020selfsupervised, tian2019contrastiveCMC, li2020prototypical}. In contrast, our simple cross-modal synthesis avoids all of these design issues and compute costs, while achieving state of the art performance in both vector and raster view downstream performance. 
\cut{We attribute this to the transformation between pixel-array and vector-sequence models of sketches being a much more challenging mapping to learn, and thus providing a stronger representation in both modalities. }


\begin{figure*}[t]
\begin{center}
  \includegraphics[width=\linewidth]{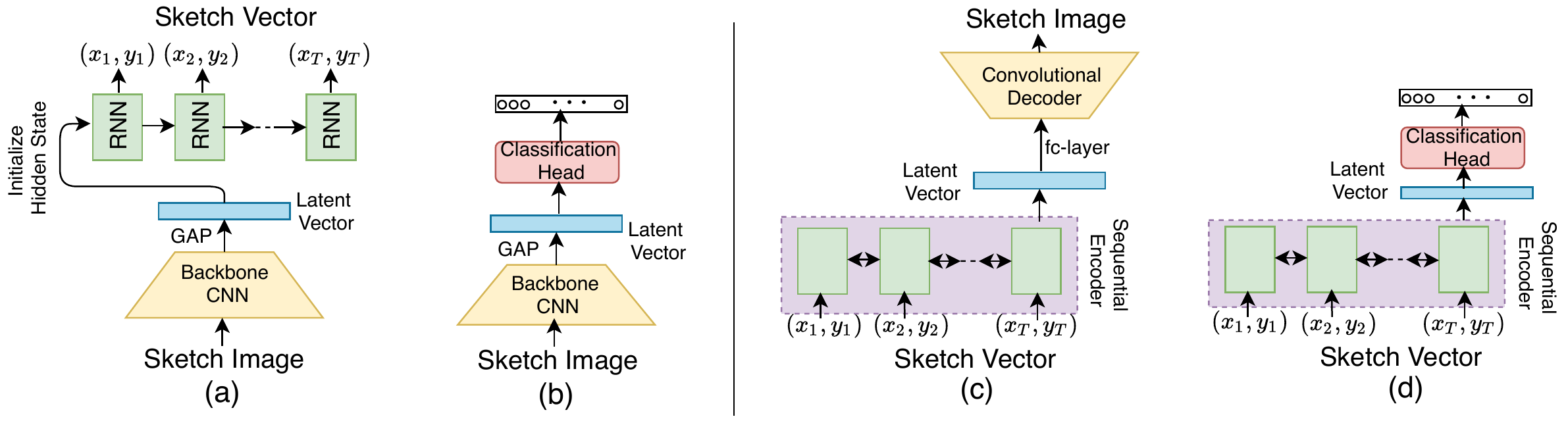}
\end{center}
\vspace{-.2in}
  \caption{Illustration of the architecture used for our self-supervised task for sketches and handwritten data (a,c), and how it can subsequently be adopted for downstream tasks (b,d). Vectorization involves translating sketch image to sketch vector (a), and the convolutional encoder used in the vectorization process acts as a feature extractor over sketch images for downstream tasks (b). On the other side, rasterization converts sketch vector to sketch image (c), and provides an encoding for vector-based recognition tasks downstream (d).}
\vspace{-.20in}
\label{fig:Fig2}
\end{figure*}

\vspace{-0.1cm}
\section{Methodology}
\vspace{-0.1cm}
\keypoint{Overview:} Our objective is to design a self-supervised learning method that can be applied over both rasterized image and vector representation of any hand drawn data (e.g., sketch or handwriting); and furthermore it should exploit this complementary information for self-supervised learning. Towards this objective, we pose the feature learning task as a cross-modal translation between image and vector space using state-of-the-art encoder-decoder architectures. In other words, the training objective is to learn a latent space for the source modality from which the corresponding sample in the target modality is predictable. Once the cross-modal translation model is trained, we can remove the decoder and use the encoder as a feature extractor for source modality data. For instance, learning to translate from image space to vector space, we obtain an encoder that can embed raster encoding of hand drawn images into a meaningful latent representation, and vice-versa. 

Touch-screen devices and stylus-pens give us easy access to hand drawn data represented in both modalities simultaneously. Our training dataset consists of N  samples $\{I_i, V_i\}_{i=1}^{N}$, where  $I \in \mathcal{I}$ and $V \in \mathcal{V}$ are rasterized image and vector representation respectively. In particular, $I$ is a spatially extended image of size $\mathbb{R}^{H \times W \times3}$, and $V$ is a sequence of pen states $(v_1, v_2, \cdots, v_T)$, where $T$ is the length of the sequence. In order to learn feature representation on image space, we learn a \emph{vectorization} operation $\mathrm{\mathcal{I} \mapsto \mathcal{V}}$. Conversely, a \emph{rasterization} operation $\mathrm{\mathcal{V} \mapsto \mathcal{I}}$ is trained to provide a vector space representation. It is important to note that we do not use any category-label for sketch data or character/word annotation for handwritten data in our feature learning process. Thus, it can be trained in a class agnostic manner without any manual labels, satisfying the criteria of self-supervised learning. 

\vspace{-0.1cm}
\subsection{Model Architecture} \label{basemodel}
\vspace{-0.1cm}
For cross-modal translation, encoder $E(\cdot)$ embeds the data from source modality into a latent representation, and decoder $D(\cdot)$ reconstructs the target modality given the latent vector. $E(\cdot)$ and $D(\cdot)$ will be designed differently for the different source and target modalities. While rasterized image space is represented by a three-channel RGB image $\mathbb{R}^{H \times W \times 3}$, we use five-element vector $v_t=(x_t, y_t, q_t^1, q_t^2, q_t^3)$ to represent pen states in stroke-level modelling. In particular, $(x_t, y_t)$ is absolute coordinate value in a normalised $H \times W$ canvas, while the last  three  elements  represent  binary  one-hot vector \cite{ha2017neural} of  three  pen-state  situations:   pen touching  the  paper,  pen  being  lifted and end  of drawing. Thus, the size of vector representation is $V\in\mathbb{R}^{T\times5}$, where T is the sequence length. 
 
\keypoint{Vectorization:} For translating an image to its sequential point coordinate equivalent, image encoder ${E_I(\cdot)}$ can be any state-of-the-art convolutional neural network \cite{kolesnikov2019revisiting} such as ResNet. To predict the sequential point coordinates, decoder $D_V(\cdot)$ could be any sequential network, e.g. RNN. In particular, given an image $I$, let the extracted convolutional feature map be $F = E_I(I) \in \mathbb{R}^{h \times w \times d}$, where $h$, $w$ and $d$ signify height, width and number of channels respectively. Applying global max pooling (GAP) to $F$ and flattening, we obtain a vector $l_I$ of size $\mathbb{R}^d$, which will be used as the representation for input image $I$ once the encoder-decoder model is trained. Next, a linear-embedding layer is used to obtain the initial hidden state of the decoder RNN as follows: $h_0 = W_hl_I+ b_h$.  The hidden state $h_t$ of decoder RNN  is updated as follows: ${h_t = RNN(h_{t-1} ; [l_I, P_{t-1}] )}$, where  $P_{t-1}$ is the last predicted point and $[\cdot]$ stands for a concatenation operation. Thereafter, a fully-connected layer is used to predict five-element vector at each time step as: $P_t = W_yh_t+ b_y$, where $P_t = (x_t, y_t, q^1_t, q^2_t, q^3_t)$ is of size $\mathbb{R}^{2+3}$, whose first two logits represent absolute coordinate $(x,y)$ and the latter three  for pen's state position $(q^1, q^2, q^3)$. We use simple mean-square error and categorical cross-entropy losses to train the absolute coordinate and pen state prediction (softmax normalised) respectively. Thus, $(\hat{x}_t, \hat{y}_t, \hat{q}^1_t, \hat{q}^2_t, \hat{q}^3_t)$ being the ground-truth coordinate at $t$-th step, the training loss is: 
\vspace{-0.3 cm}
\begin{equation}
\begin{aligned}
L_{I \rightarrow V} =  &\frac{1}{T}\sum_{t=1}^{T} \left \| \hat{x}_t - {x}_t \right \|_2 +  \left \| \hat{y}_t - {y}_t\right \|_2 \\[-4pt] 
- &\frac{1}{T}\sum_{t=1}^{T}\sum_{i=1}^{3}\hat{q^i_t} \log \Big(\frac{\exp({q^i_t})}{\sum_{j=1}^{3}\exp({q^j_t})}\Big)
\end{aligned}
\end{equation}

\keypoint{Rasterization:} To translate a sequence of point coordinates $V$ to its equivalent image representation $I$, any sequential network such as RNN \cite{collomosse2019livesketch, graves2008novel} or Transformer \cite{sketchbert2020}, could be used as the encoder $E_V(\cdot)$, and we experiment with both. For RNN-like architectures \cite{ha2017neural}, we feed the five elements vector $v_t$ at every time step, and take the hidden state of final time step as the encoded latent representation. For Transformer like encoders, we take input via a trainable linear layer to convert each five element vector to the Transformer's model dimension. Additionally, we prepend a learnable embedding to the input sequence, similar to BERT's \emph{class token} \cite{devlin2018bert}, whose state at the output acts as the encoded latent representation.  Finally, the encoder latent representation $l_V\in\mathbb{R}^d$ is fed via a fully-connected layer to a standard convolutional decoder $D_I(\cdot)$. $D_I(\cdot)$ consists of series of fractionally-strided convolutional layers \cite{isola2017image} to up-sample the spatial size to $H \times W$ at the output. \cut{Given the corresponding rasterized image $\hat{I}$,  and the generated output from the convolutional decoder be  $I$, we use mean-square error as the training objective:} Given the vector $V$ and raster $I$ data pairs, we  use mean-square error as the training objective:

\vspace{-0.6cm}
\begin{equation}
L_{V \rightarrow I} = - \mathbb{E}_{(I, V) \sim \mathcal{(I,V)}}  \left \| I -  D_I(E_V(V))\right \|_2 
\end{equation}
\vspace{-0.5cm}

We remark that due to the well known regression to mean problem \cite{mathieu2015deep}, the generated images are indeed blurry. Adding an adversarial loss \cite{isola2017image} does not give any improvement in our representation learning task, and sometimes leads to worse results due to mode collapse issue in adversarial learning. However, synthesising realistic images is not our goal in this work. Rather, it is a pretext task for learning latent representations for vector sequence inputs. 

\subsection{Application of Learned Representation}

We apply our self-supervised learning method on both sketch \cite{ha2017neural, eitz2012humans} and handwriting data \cite{IamDataset}, as both can be represented in image and vector space. 

\keypoint{Sketch Analysis:} We use sketch-recognition \cite{yu2016sketchAnet} and sketch-retrieval \cite{xu2018sketchmate, sketchbert2020} as downstream tasks to evaluate the quality of learned latent representation obtained by our self-supervised pre-training. For both classification and retrieval, we evaluate performance with both sketch image and sketch vector representations using vectorization and rasterization as pre-training task respectively. For classification, we simply apply a fully-connected layer with softmax on the extracted latent representation from pre-trained encoder. For retrieval, we could use the latent representation itself. However we find it helpful to project the latent feature through a fully connected layer into $256$ dimensional embedding space, and optimise the model through triplet loss \cite{dey2019doodle, bhunia2020sketch}. Along with triplet loss, that minimises intra-class distance while maximising inter-class distance, we also apply a classification loss through a linear layer \cite{collomosse2019livesketch} to further aid the retrieval learning framework.  

\keypoint{Handwriting Recognition:} 
As handwriting recognition \cite{bhunia2019handwriting} is a (character) sequence task oriented at decoding a whole word from an image\cut{ (`sketching' the word during pre-training, and recognising the letters as the downstream task)}, we use a slightly modified vectorization encoder $E_I(\cdot)$ for offline/image recognition compared to the sketch tasks. Following \cite{shi2018aster}, the (word) image encoder extends a conventional ResNet architecture with a 2-layer BLSTM image feature encoder before producing a final state that provides the latent vector for the input image. This feature is then fed to a sequential decoder to `sketch' the word during cross-modal self-supervised pre-training, and to a recognition model to recognise the word in the downstream task. In contrast, the rasterization encoder $E_V(\cdot)$ is defined similarly as for the sketch tasks. For the downstream task, after encoding either vector and raster inputs, we follow \cite{oord2018representation} in using an attentional decoder \cite{shi2018aster, zhang2019sequence} to recognize the word by predicting the characters sequentially. This decoder module consists of a BLSTM layer followed by a GRU layer that predicts the characters.

\vspace{-0.2cm}
\section{Experiments}\label{sec:experiments}
\vspace{-0.15cm}
\noindent \textbf{Datasets:} For sketches, we use the standard QuickDraw \cite{ha2017neural} and TU-Berlin \cite{eitz2012humans} datasets for evaluation as they contain both raster and vector image representations. QuickDraw contains 50 million sketches from 345 classes. We use the split from \cite{ha2017neural} where each class has 70K training samples, 2.5K validation, and 2.5K test samples. Meanwhile, TU-Berlin comprises of 250 object categories with 80 sketches in each category. We apply the Ramer-Douglas-Peucker (RDP) algorithm to simplify the sketches. For handwriting we use IAM offline and online  datasets \cite{IamDataset}: The offline set  contains 115,320 word images, while the online set contains point coordinate representation of $13,049$ lines of  handwriting.  We pre-process line-level online data to segment it into $70,648$ valid words, and use synthetic rasterization to create training data for vector and raster views.

\noindent \textbf{Implementation Details:} We implemented our framework in PyTorch \cite{paszke2017automatic} and conducted experiments on a 11 GB Nvidia RTX 2080-Ti GPU. 
While a GRU decoder of hidden state size $512$ is used in all the vectorization process, we use convolutional decoder from \cite{isola2017image}  in the rasterization process. Following a recent self-supervised study analysis \cite{kolesnikov2019revisiting, grill2020bootstrap}, we use ResNet50 as the CNN encode images, unless otherwise mentioned. For vector sketch recognition, we use a  Transformer \cite{vaswani2017attention} encoder with 8 layers, hidden state size 768, MLP size 2048, and 12 heads.  For offline handwritten images, the encoder architecture is taken from \cite{shi2018aster} and comprises a ResNet like convolutional architecture followed by a 2 layers BLSTM. For online handwriting, we feed 5-element vectors at every time step of a 4-layers stacked BLSTM \cite{carbune2020fast} with hidden state size 512.  We use Adam optimiser with learning 0.0001 and batch size of 64 for all experiments.

\setlength{\tabcolsep}{3pt}
\begin{table*}[!hbt]
    \centering
    \scriptsize
    \caption{Linear model evaluation of fixed pre-trained features. ResNet50 for image space and Transformer for vector space inputs.}
    \begin{tabular}{r|cccccccc|cccccccc}
        \hline \hline
         & \multicolumn{8}{c|}{Recognition} & \multicolumn{8}{c}{Retrieval} \\
        \cline{2-17}
         & \multicolumn{4}{c}{Image Space} & \multicolumn{4}{c|}{Vector Space} & \multicolumn{4}{c}{Image Space} & \multicolumn{4}{c}{Vector Space} \\
        \cline{2-17}
         & \multicolumn{2}{c}{QuickDraw} & \multicolumn{2}{c}{TU-Berlin} & \multicolumn{2}{c}{QuickDraw} & \multicolumn{2}{c|}{TU-Berlin} & \multicolumn{2}{c}{QuickDraw} & \multicolumn{2}{c}{TU-Berlin} & \multicolumn{2}{c}{QuickDraw} & \multicolumn{2}{c}{TU-Berlin} \\
         & Top-1 & Top-5 & Top-1 & Top-5 & Top-1 & Top-5 & Top-1 & Top-5 & A@T1 & mAP@t10 & A@T1 & mAP@t10 & A@T1 & mAP@T10 & A@T1 & mAP@T10 \\
        \hline
        Supervised & 76.1\% & 91.3\% & 78.6\% & 90.1\% & 73.5\% & 90.1\% & 62.9\% & 80.7\% & 62.3\% & 69.4\% & 69.1\% & 74.7\% & 58.5\% & 77.1\% & 50.2\% & 67.4\% \\

        Random & 15.5\% & 26.2\% & 18.4\% & 29.3\% & 12.7\% & 23.6\% & 9.6\% & 19.4\% & 10.6\% & 21.3\% & 13.4\% & 26.7\% & 9.8\% & 21.5\% & 9.2\% & 17.6\% \\
        \hdashline 
        Context \cite{doersch2015unsupervised} &  44.6\% & 69.2\% & 43.3\% & 67.5\% & - & - & - & - & 30.7\% & 34.9\% & 28.4\% & 32.7\% & - & -& - & - \\
        Auto-Encoder \cite{kingma2013auto} &  26.4\% & 48.1\% & 22.6\% & 47.5\% & - & - & - & - & 16.4\% & 24.4\% & 15.3\% & 20.4\% & - & -& - & - \\
        Jigsaw \cite{noroozi2016unsupervised} &  46.9\% & 71.5\% & 45.7\% & 69.8\% & - & - & - & - & 31.6\% & 38.9\% & 30.6\% & 35.4\% & - & -& - & - \\
       Rotation \cite{gidaris2018unsupervised} &  53.5\% & 78.7\% & 51.2\% & 77.1\% & - & - & - & - & 37.5\% & 45.1\% & 36.4\% & 41.8\% & - & -& - & - \\
     Deep Cluster \cite{caron2018deep} & 39.4\% & 62.7\% & 38.7\% & 60.2\% & - & - & - & - & 29.2\% & 36.8\% & 27.3\% & 31.9\% & - & -& - & - \\
    MoCo \cite{he2020momentum} &  65.7\% & 85.1\% & 64.3\% & 82.8\% & - & - & - & - & 42.5\% & 46.8\% & 42.5\% & 46.9\% & - & -& - & - \\
   SimCLR \cite{chen2020simple} &  65.5\% & 85.1\% & 64.3\% & 82.9\% & - & - & - & - & 43.3\% & 50.7\% & 41.5\% & 46.7\% & - & -& - & - \\
    BYOL \cite{grill2020bootstrap} &  66.8\% & 85.8\% & 65.7\% & 83.7\% & - & - & - & - & 45.4\% & 52.5\% & 43.8\% & 49.1\% & - & -& - & - \\
Sketch-BERT \cite{sketchbert2020} & - & - & - & - & 65.6\% & 85.3\% & 52.9\% & 78.1\% & - & - & - & - & 48.9\% & 68.1\% & 40.7\% & 58.8\% \\
CMC \cite{tian2019contrastiveCMC}  & 63.6\% & 83.9\% & 61.7\% & 81.3\% & 61.2\% & 81.5\% & 51.4\% & 77.5\% & 40.6\% & 45.8\% & 38.5\% & 43.3\% & 45.2\% & 66.7\% & 40.3\% & 58.2\% \\
CPC \cite{oord2018representation}  & 54.3\% & 79.0\% & 52.9\% & 77.9\% & 59.3\% & 81.3\% & 50.5\% & 76.6\% & 37.9\% & 43.1\% & 36.4\% & 40.9\% & 43.1\% & 63.6\% & 39.3\% & 57.9\% \\
Ours-(L)  & 71.9\% & 89.7\% & 70.6\% & 85.9\% & 67.2\% & 86.5\% & 55.6\% & 79.4\% & 52.3\% & 59.5\% & 47.7\% & 59.1\% & 49.5\% & 68.9\% & 42.1\% & 59.6\% \\
        \hline
    \end{tabular}
    \label{tab:my_label1}
    \vspace{-0.3cm}
\end{table*}

\setlength{\tabcolsep}{3pt}
\begin{table*}[]
    \centering
    \scriptsize
    \caption{Semi-supervised fine-tuning using $1\%$ and $10\%$ labelled training data on QuickDraw.}
    \begin{tabular}{r|cccccccc|cccccccc}
        \hline \hline
         & \multicolumn{8}{c|}{Recognition} & \multicolumn{8}{c}{Retrieval} \\
        \cline{2-17}
         & \multicolumn{4}{c}{Image Space} & \multicolumn{4}{c|}{Vector Space} & \multicolumn{4}{c}{Image Space} & \multicolumn{4}{c}{Vector Space} \\
        \cline{2-17}
         & \multicolumn{2}{c}{1\% Training} & \multicolumn{2}{c}{10\% Training} & \multicolumn{2}{c}{1\% Training} & \multicolumn{2}{c|}{10\% Training} & \multicolumn{2}{c}{1\% Training} & \multicolumn{2}{c}{10\% Training} & \multicolumn{2}{c}{1\% Training} & \multicolumn{2}{c}{10\% Training} \\
         & Top-1 & Top-5 & Top-1 & Top-5 & Top-1 & Top-5 & Top-1 & Top-5 & A@T1 & mAP@t10 & A@T1 & mAP@t10 & A@T1 & mAP@T10 & A@T1 & mAP@T10 \\
        \hline
        Supervised & 25.1\% & 47.3\% & 55.4\% & 79.0\% & 17.3\% & 37.5\% & 43.9\% & 65.9\% & 13.4\% & 34.3\% & 43.9\% & 63.7\% & 9.1\% & 29.0\% & 41.0\% & 60.8\% \\
       \hdashline
        Context \cite{doersch2015unsupervised} &  33.9\% & 55.8\% & 56.8\% & 80.5\% & - & - & - & - & 24.4\% & 30.5\% & 42.6\% & 48.4\% & - & -& - & - \\
        Auto-Encoder  \cite{kingma2013auto} &  21.5\% & 40.7\% & 45.1\% & 70.6\% & - & - & - & - & 15.2\% & 21.4\% & 32.7\% & 37.6\% & - & -& - & - \\
        Jigsaw \cite{noroozi2016unsupervised} &  36.5\% & 57.4\% & 57.4\% & 80.3\% & - & - & - & - & 27.7\% & 35.2\% & 44.7\% & 51.2\% & - & -& - & - \\
       Rotation  \cite{gidaris2018unsupervised}  &  38.8\% & 59.1\% & 59.6\% & 80.7\% & - & - & - & - & 28.4\% & 35.2\% & 44.7\% & 51.8\% & - & -& - & - \\
     Deep Cluster \cite{caron2018deep}  & 32.2\% & 54.5\% & 54.7\% & 79.2\% & - & - & - & - & 24.4\% & 31.2\% & 43.6\% & 47.7\% & - & -& - & - \\
    MoCo \cite{he2020momentum}  &  46.0\% & 70.5\% & 62.2\% & 83.7\% & - & - & - & - & 35.9\% & 43.1\% & 52.7\% & 57.4\% & - & -& - & - \\
   SimCLR  \cite{chen2020simple} &  46.1\% & 70.5\% & 62.1\% & 83.6\% & - & - & - & - & 35.1\% & 42.7\% & 52.3\% & 57.4\% & - & -& - & - \\
    BYOL  \cite{grill2020bootstrap} &  47.3\% & 72.0\% & 62.7\% & 84.1\% & - & - & - & - & 36.5\% & 43.0\% & 52.8\% & 59.8\% & - & -& - & - \\
Sketch-BERT \cite{sketchbert2020}  & - & - & - & - & 45.1\% & 69.8\% & 62.4\% & 81.7\% & - & - & - & -& 36.5\% & 60.0\% & 52.9\% & 72.9\% \\
CMC \cite{tian2019contrastiveCMC}  & 44.6\% & 68.2\% & 61.7\% & 82.7\% & 44.6\% & 68.4\% & 61.7\% & 81.6\% & 34.7\% & 41.9\% & 51.1\% & 57.4\% & 35.4\% &  58.1\% & 52.6\% & 72.8\% \\
CPC \cite{oord2018representation}  & 40.6\% & 65.7\% & 60.7\% & 81.9\% & 43.5\% & 67.7\% & 61.6\% & 81.7\% & 33.4\% & 40.1\% & 50.5\% & 57.7\% & 34.1\% & 56.6\% & 52.3\% & 72.8\% \\
Ours  & 51.2\% & 76.4\% & 65.6\% & 85.2\% & 46.8\% & 70.9\% & 63.2\% & 83.9\% & 38.6\% & 45.6\% & 60.4\% & 81.4\% & 37.1\% & 61.5\% & 53.2\% & 74.3\% \\
        \hline
    \end{tabular}
    \label{tab:my_label2}
    \vspace{-0.4cm}
\end{table*}
 
\setlength{\tabcolsep}{6pt}
\begin{table}[h]
    \centering
    \footnotesize
    \caption{Accuracy on QuickDraw dataset with linear classifier trained on representation from various depth within the network.}
    \begin{tabular}{cccccc}
        \hline \hline
        Method & Block1 & Block2 & Block3 & Block4 & Pre-logits\\
        \hline
         Supervised & 7.0\% & 14.9\% & 35.6\% & 72.5\% & 76.1\% \\
         \hdashline
         Jigsaw \cite{noroozi2016unsupervised} & 4.2\% & 8.1\% & 26.8\% & 39.9\% & 46.9\% \\
         Rotation  \cite{gidaris2018unsupervised} & 5.2\% & 11.2\% & 27.5\% & 45.4\% & 53.5\% \\
         Deep Cluster \cite{caron2018deep}  & 4.1\% & 8.6\% & 19.6\% & 33.4\% & 39.4\% \\
       CMC \cite{tian2019contrastiveCMC}  & 6.1\% & 11.9\% & 29.7\% & 56.7\% & 63.6\% \\
         MoCo \cite{he2020momentum}  & 7.7\% & 13.5\% & 31.6\% & 60.3\% & 65.7\%  \\
         SimCLR \cite{chen2020simple} & 6.7\% & 12.6\% & 32.0\% & 59.5\% & 65.5\% \\
         BYOL  \cite{grill2020bootstrap} & 9.0\% & 15.1\% & 32.2\% & 61.2\% & 66.8\% \\
         Ours & 10.1 & 15.2\% & 34.4\% & 67.5\% & 71.9\% \\
        \hline
    \end{tabular}
    \label{tab:my_label3}
\vspace{-0.3cm}
\end{table}

\noindent \textbf{Evaluation Metrics:}  For sketch recognition, Top-1 and Top-5 accuracy is used, and for category level sketch-retrieval, we employ Acc@top1 and mAP@top10 as the evaluation metric. For handwriting recognition, we use Word Recognition Accuracy following \cite{bhunia2019handwriting}.

\noindent \textbf{Competitors:}  We compare with existing self-supervised learning methods that involve pre-text task like context prediction \cite{doersch2015unsupervised},  Auto-Encoding \cite{kingma2013auto}, jigsaw solving \cite{noroozi2016unsupervised}, rotation prediction \cite{gidaris2018unsupervised}. Clustering based representation learning Deep Cluster  \cite{caron2018deep} is also validated on sketch datasets. Furthermore, we compare with three state-of-the-art contrastive learning based self-supervised learning methods, namely, SimCLR \cite{chen2020simple}, MoCo \cite{he2020momentum}, and BYOL \cite{grill2020bootstrap}. While these self-supervised learning methods are oriented at RGB photos rather than sketch or handwritten data, we also compare with Sketch-Bert \cite{sketchbert2020} as the only work employing self-supervised learning on vector sketches. We note that self-supervised methods designed for image data can not be used off-the-shelf for vector sketches. The only exception is Contrastive Predictive Coding (CPC) \cite{oord2018representation}  which has been used to handle both images and sequential data (e.g. speech signals). Finally, we compare with a state of the art multi-modal self-supervised method Contrastive Multi-view Coding (CMC) \cite{tian2019contrastiveCMC}, which performs contrastive learning of (mis)matching instances across modalities. Here, we use the same encoder for raster sketch and vector sketch like ours for a fair comparison. 

\subsection{Results on Sketch Representation Learning}\label{sec:perf}
\begin{figure*}[t]
        \centering
		\includegraphics[height=2.7cm, width=0.24\linewidth]{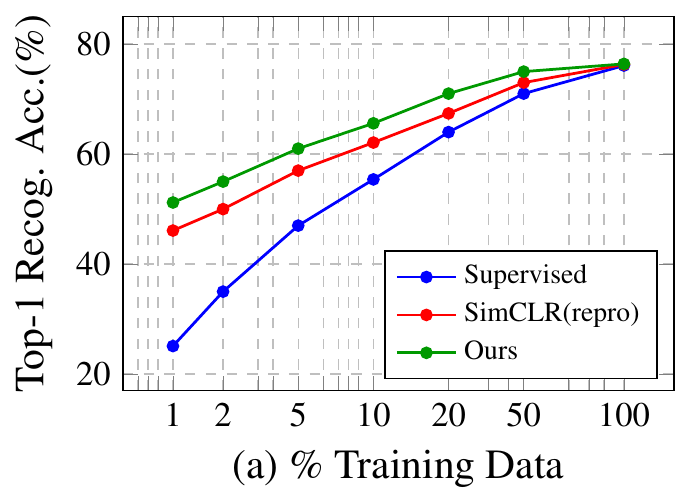}
 		\includegraphics[height=2.7cm, width=0.24\linewidth]{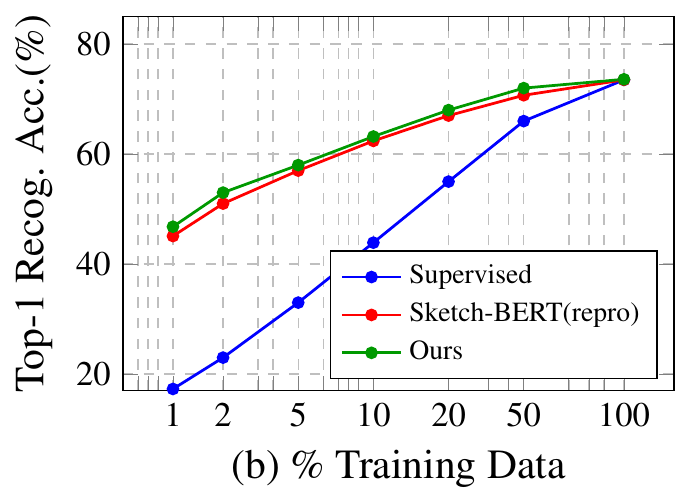}
		\includegraphics[height=2.7cm, width=0.24\linewidth]{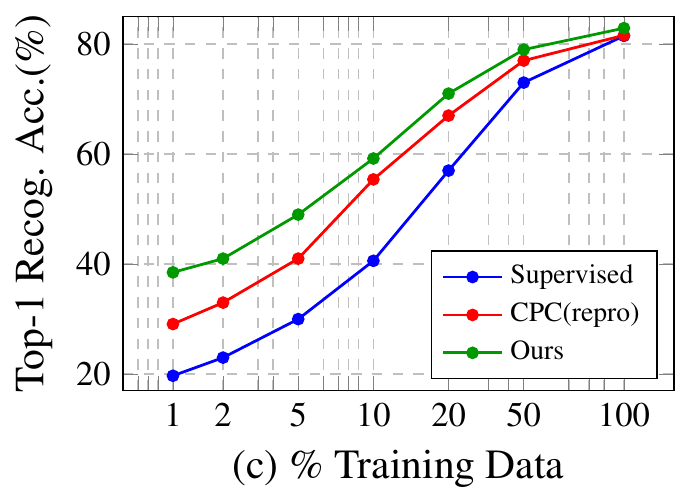}
		\includegraphics[height=2.7cm, width=0.24\linewidth]{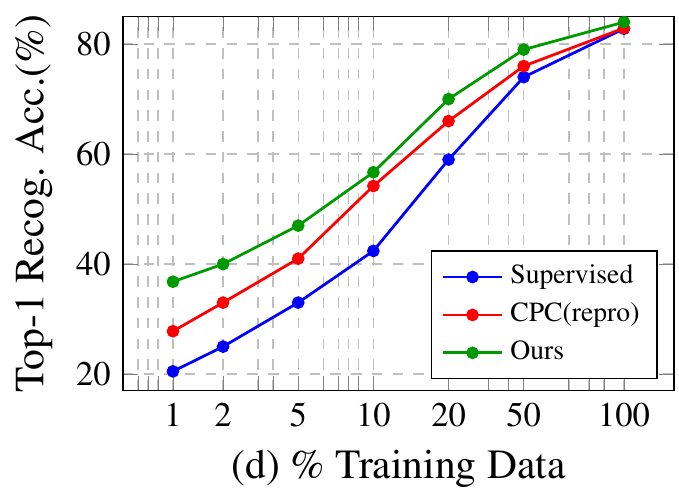} \\ \hspace{0.25cm}
		\includegraphics[height=2.7cm, width=0.24\linewidth]{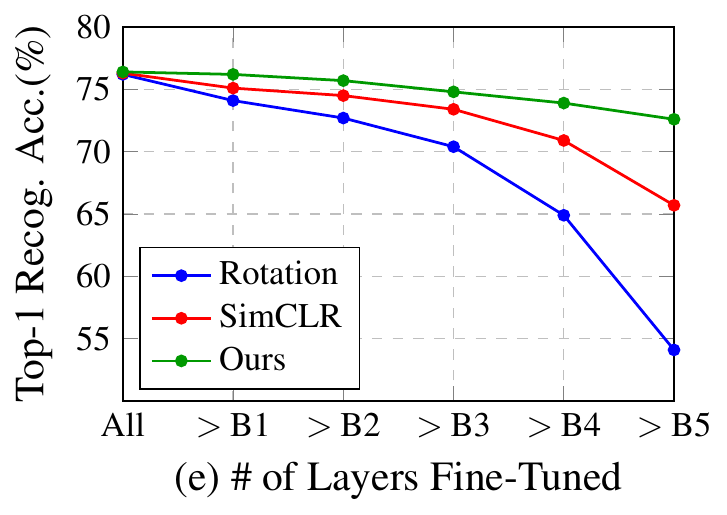}
		\includegraphics[height=2.7cm, width=0.24\linewidth]{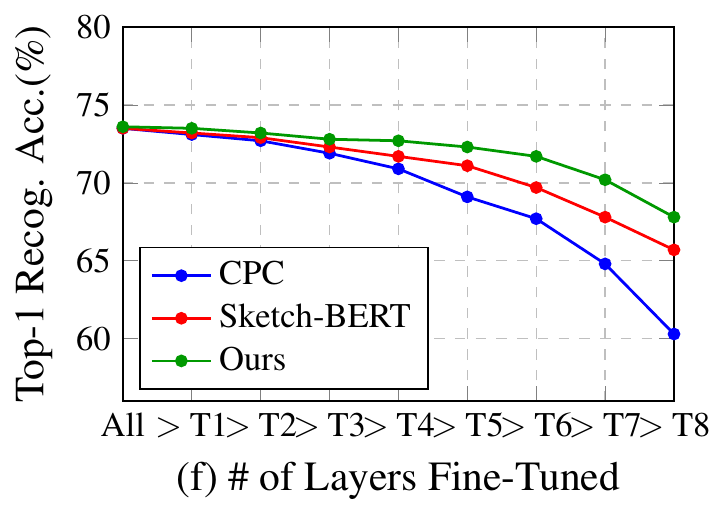}
		\includegraphics[height=2.7cm, width=0.24\linewidth]{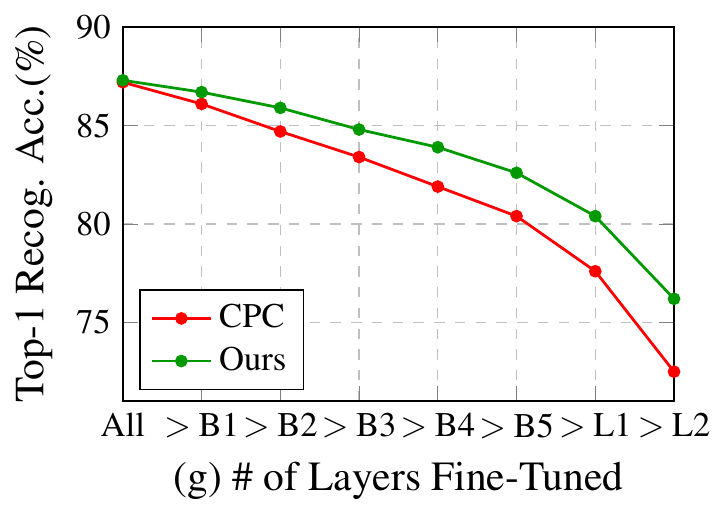}
		\includegraphics[height=2.7cm, width=0.24\linewidth]{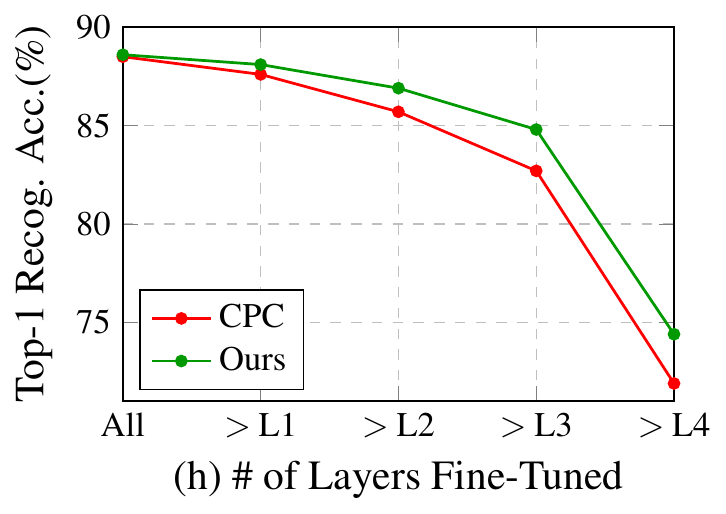}
	\vspace{-0.05in}
	\caption{Performance at varying training data size for (a) sketch image classification (b) sketch vector classification on Quick-Draw, and (c) offline handwritten image recognition (d) online handwriting recognition, respectively. In the same order, comparative performance is shown through fine-tuning different number of layers: (e) sketch image encoder uses ResNet-50 having 5-convolutional blocks. (f) 8-layers stacked Transformer is used for sketch vector encoder. (g) ResNet-like convolutional encoder (having 5 blocks) followed by 2 layers BLSTM employs offline word image encoder (h) 4 layers stacked BLSTM is used for encoder online word images.`$\mathrm{>X}$' represents all layers above $X$ are fine-tuned. 
	\label{plots}
\vspace{0.1cm}
 }
	\label{fig:data_layers}
	\vspace{-.1in}
\end{figure*}

\vspace{-0.1cm}
\keypoint{Sketch Recognition:}  Following the traditional protocol of evaluation for self-supervised learning \cite{chen2020simple, grill2020bootstrap}, we first evaluate our representations by training a linear classifier on the top of frozen representation. We report the recognition accuracy in  Table(left) \ref{tab:my_label1}. On QuickDraw, Top-1 accuracy of $71.9\%$ and $67.2\%$ is obtained for sketch images and sketch vectors respectively, approaching the supervised counterparts of $76.1\%$ and $73.5\%$. For TU-Berlin accuracies of $70.6\%$ and $55.6\%$ also approach the supervised figures $78.6\%$ and $62.9\%$. The gap with supervised method is larger for TU-Berlin dataset because of having less data compared to QuickDraw dataset. Interestingly, the performance over image level data is comparatively better than using sketch vectors.  

We next evaluate the semi-supervised setup, where we fine-tune the whole network using smaller subset of training data, $1\%$ and $10\%$. Our self-supervised methods helps to learn good initialization  such that in this low data regime, ours is significantly better than its supervised counter part as shown in Table \ref{tab:my_label2}. Finally, we evaluate the learned features from various depths of our convolutional encoder for sketch raster image classification in Table \ref{tab:my_label3}. Overall, our close competitors are contrastive learning based family of self-supervised methods, e.g. SimCLR, BYOL, MoCo. We attribute the superiority of our method over other self-supervised methods, on the sketch dataset, to the task-design that exploits the intrinsic dual representation of sketch data.

\begin{figure*}[]
	\begin{center}
		\includegraphics[ width=1\linewidth]{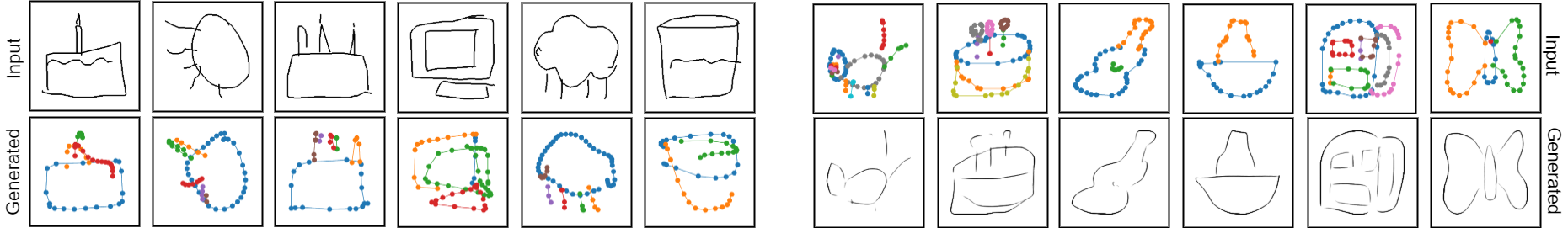}
	\end{center}
	\vspace{-0.15in}
	\caption{Qualitative results showing generated cross-modal translation. (a) Vectorization: raster sketch image to vector sketch, (b) Rasterization: vector sketch to raster sketch image.}
	\label{fig:Graph1_Full}
	\vspace{-.21in}
\end{figure*}

\begin{figure*}[]
	\begin{center}
		\includegraphics[ width=1\linewidth]{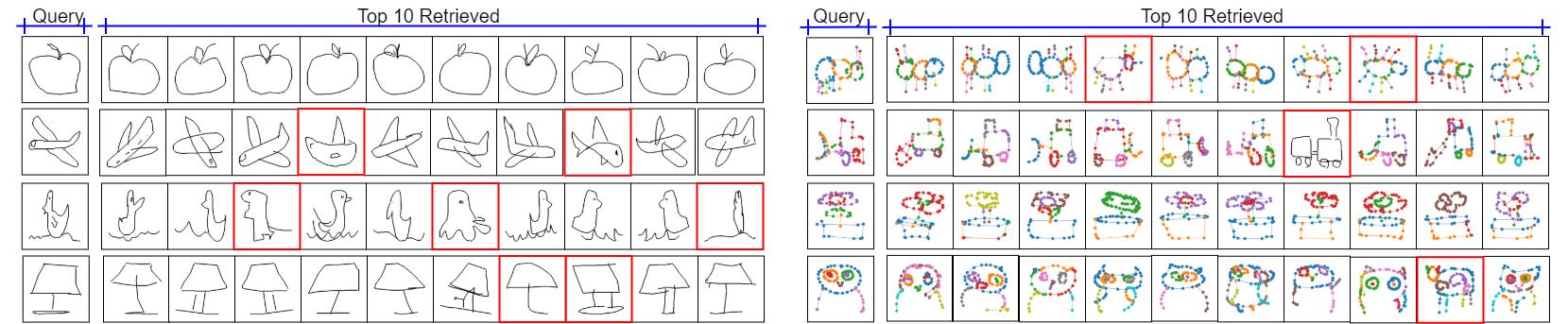}
	\end{center}
	\vspace{-0.15in}
	\caption{Qualitative retrieved results on (a)  raster sketch images (via vectorization task) (b) vector sketches (via rasterization task) using  pre-trained latent feature. Red denotes false positive cases.}
	\label{fig:Graph2_Full}
	\vspace{-.18in}
\end{figure*}

\keypoint{Sketch Retrieval:} For sketch retrieval, first we use the extracted latent feature from pre-trained self-supervised network for triplet metric learning of an additional linear embedding layer. From the retrieval performance in Table~\ref{tab:my_label1} (right) we see a relative performance between the methods that is similar to the previous sketch classification experiments. However, the retrieval performance using the fixed self-supervised latent feature is $9-10\%$ below the supervised version.  In the semi-supervised experiment, we fine-tune the complete model including  linear layer and the pre-trained feature extractor using $1\%$ and $10\%$ of the training data respectively.  Table \ref{tab:my_label2} shows that our self-supervised method has a clear edge over supervised counter part in this low data regime. Qualitative cross-modal generated and retrieved results are shown in Figure~\ref{fig:Graph1_Full} \& \ref{fig:Graph2_Full}, respectively.

\setlength{\tabcolsep}{6pt}
\begin{table}[h]
 \scriptsize
    \centering
    \caption{Handwriting recognition using feature extracted from fixed pre-trained encoder.}
    \begin{tabular}{ccccc}
        \hline\hline
         & \multicolumn{2}{c}{Offline} & \multicolumn{2}{c}{Online} \\
        \cline{2-5}
         & Lexicon & No Lexicon &  Lexicon & No Lexicon  \\
        \hline
        Supervised \cite{shi2018aster} & 87.1\% & 81.5\% & 88.4\% & 82.8\% \\
        Random & 10.4\% & 6.3\% & 7.4\% & 4.9\% \\
            \hdashline
        CPC \cite{oord2018representation} & 72.2\% & 63.7\% & 71.5\% & 62.8\% \\
        Ours & 75.4\% & 68.6\% & 73.1\% & 66.9\% \\
        \hline
    \end{tabular}
    \label{tab:my_label4}
    \vspace{-0.2cm}
\end{table}

\setlength{\tabcolsep}{6pt}
\begin{table}[h]
 \scriptsize
    \centering
    \caption{Handwriting recognition under semi-supervised setup.}
    \begin{tabular}{ccccc}
        \hline\hline
         & \multicolumn{2}{c}{Offline} & \multicolumn{2}{c}{Online} \\
        \cline{2-5}
         & 1\% Training & 10\% Training & 1\% Training & 10\% Training \\
        \hline
        Supervised \cite{shi2018aster} & 19.7\% & 40.6\% & 20.5\% & 42.4\% \\
        \hdashline
        CPC \cite{oord2018representation} & 29.1\% & 55.4\% & 27.8\% & 54.2\% \\
        Ours & 38.5\% & 59.2\% & 36.8\% & 56.7\% \\
        \hline
    \end{tabular}
    \label{tab:my_label5}
  \vspace{-0.5cm}
\end{table}

\vspace{-0.1cm}
\subsection{Results on Handwriting Recognition}\label{htr}
\vspace{-0.1cm}
To the best of our knowledge, there has been no work applying self-supervised learning to handwritten data. We compare our self-supervised  method with CPC which can handle sequential data. In Table~\ref{tab:my_label4}, we use the extracted frozen sequential feature from each encoder to train an attentional decoder based text recognition network. We see that our \modelName{} surpasses CPC, but both methods do not match supervised performance. In a semi-supervised setup (Table \ref{tab:my_label5}), we add an attentional decoder and fine-tune the whole pipeline using $1\%$ and $10\%$ training data respectively. In this case, the self-supervised methods achieve a significant margin over the supervised alternative. Furthermore, we observe that initialising the network (both offline and online) with weights pre-trained on our self-supervised setup, followed by training (supervised) the entire pipeline, yields higher results than initialising with random weights, by a margin of $1.4\%$ and $1.2\%$, under the same experimental setup. This concludes that our smart pre-training strategy is a better option, instead of training handwriting recognition network from scratch.


\vspace{-0.2cm}
\setlength{\tabcolsep}{3pt}
\begin{table}[h]
    \centering
    \footnotesize
    \caption{Ablative study (Top-1 accuracy) on architectural design using QuickDraw. (V)ectorization and (R)asterization indicate representation learning on image and vector space, respectively.}
    \begin{tabular}{lcc}
        \hline \hline
         Ablation Experiment & Image Space & Vector Space \\
        \hline
         (a) Absolute coordinate in the decoding (V): & 71.9\% & -- \\
         (b) Offset coordinate in the decoding (V) & 69.5\% & -- \\
         (c) Absolute coordinate in the encoding (R): & -- & 67.2\% \\
         (d) Offset coordinate in the encoding (R) & -- & 67.1\% \\
         (e) LSTM decoder (V) : & 70.7\% & -- \\
         (f) GRU decoder (V) : & 71.9\% & -- \\
         (g) Transformer decoder (V) : & 68.6\% & -- \\
         (h) LSTM encoder (R) : & -- & 66.7\% \\
         (i) GRU encoder (R) : & -- & 66.1\% \\
         (j) Transformer encoder (R) : & -- & 67.2\% \\
         (k) Two-way Translation (V+R) : & 70.3\% & 66.1\% \\
         (l) Attentional Decoder (V) : & 68.0\% & -- \\
        \hline
    \end{tabular}
    \label{tab:architechture}
    \vspace{-0.37cm}
\end{table}

\vspace{-0.2cm}
\subsection{Ablative Study}\label{abla}
\vspace{-0.15cm}

\begin{figure}[]
	\begin{center}
		\includegraphics[height=3.8cm, width=1\linewidth]{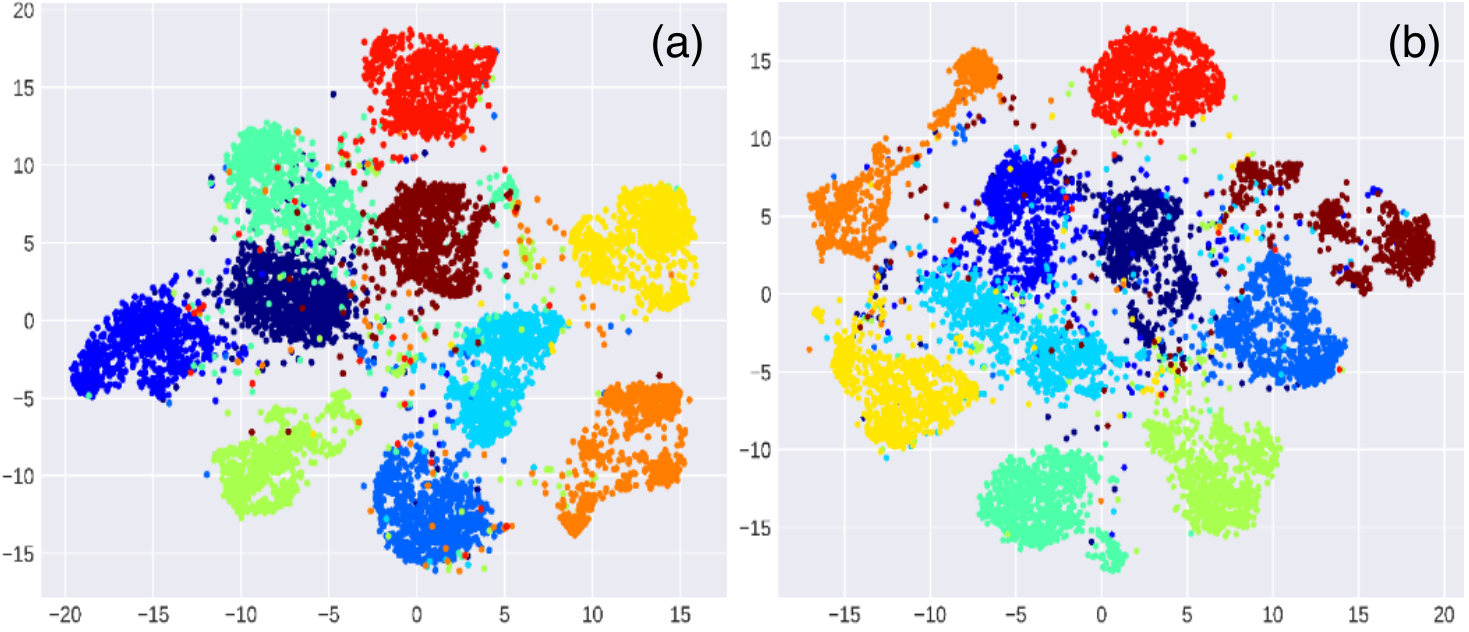}
	\end{center}
	\vspace{-0.15in}
	\caption{T-SNE Plots on features extracted by our self-supervised method (a) vectorization (sketch images) (b) rasterization (sketch vectors) for 10 QuickDraw classes.}
	\label{fig:Graph3_Full}
	\vspace{-.07in}
\end{figure}

\keypoint{Data Volume and Layer Dependence:} 
Performance under varying amounts of training data is shown in Figure \ref{fig:data_layers} for both sketch classification and handwriting recognition. We also evaluate performance as a function of number of trained/frozen layers during fine-tuning. We can see that \modelName{} performs favorably to state of the art alternatives SimCLR, SketchBERT, and CPC -- especially in the low data, or few tuneable layers regimes. 

\noindent \textbf{Architectural Insights:} We perform a thorough ablative study to provide insights on our architecture design choices using the sketch recognition task in Table~\ref{tab:architechture}. (i) In the sketch image to sketch vector translation process, using  \emph{absolute coordinate}  is found to give better representative feature for sketch images over using offset coordinate \cite{ha2017neural} values. However, for representation learning over vector sketches, we did not notice any significant difference in performance, provided the absolute coordinates are normalised.  (ii) We found absolute coordinate with regression loss gives better performance than using offset with log-likelihood loss as used in \cite{song2018learning}. (iii) We use deterministic cross-modal encoder-decoder architecture since VAE-based design \cite{song2018learning} reduces the performance. (iv) We also compare with different sequential decoders in the vectorization process, e.g. LSTM, GRU, and Transformer. Empirically, GRU is found to work better than others. (v) For sketch classification on vector space, we also compare with  LSTM, GRU, and Transformer encoder architecture respectively, with Transformer giving optimum results. (vi) Another intuitive alternative could be to use two-way cross-modal translation using additional source-to-source and target-to-target decoder, however, we experience performance drop. (vii) We also add an attentional block for sequential decoding in vectorization process that leads to a drop in performance by $3.9\%$. We conjecture that adding attention gives a shortcut connection to the convolutional feature map, and the sequential task becomes comparatively easy, which is why the self-supervised pre-training fails to learn global semantic representation for classification.  

\noindent \textbf{Cross-category Generalisation:} One major objective of unsupervised representation learning is to learn feature representation that can generalise to other categories as well. Thus, we split 345 QuickDraw classes into two random disjoint set \cite{dey2019doodle} of $265$ and $80$ for self-supervised training and evaluation, respectively. Model trained using our self-supervised task, is further evaluated on unseen classes (Table \ref{tab:cross-category}) using a linear classifier on extracted frozen feature. We obtain a top-1 accuracy of $65.1\%$ and  $58.4\%$ on sketch images and vectors, respectively, compared to  $71.9\%$ and  $67.2\%$ while using all classes in the self-supervised pre-training. Under same setting, SimCLR is limited to $53.6\%$ for sketch image classification. This confirms a significant extent of generalizable feature learning through our self-supervised task over sketch data.  

\vspace{-0.25cm} 
\setlength{\tabcolsep}{10pt}
\begin{table}[!h]
    \centering
    \footnotesize
    \caption{Cross-category recognition accuracy on QuickDraw.}
    \begin{tabular}{c|cc|cc}
        \hline \hline
         & \multicolumn{2}{c|}{Image Space} & \multicolumn{2}{c}{Vector Space} \\
        \cline{2-5}
         & Top-1 & Top-5 & Top-1 & Top-5 \\
        \hline
        MoCo \cite{he2020momentum} & 53.4\% & 77.6\% & -- & -- \\
        SimCLR \cite{chen2020simple} & 53.6\% & 77.6\% & -- & -- \\
        CPC \cite{oord2018representation} & 46.8\% & 71.3\% & 48.1\% & 73.3\% \\
        Ours & 65.1\% & 85.6\% & 58.4\% & 81.2\% \\
        \hline
    \end{tabular}
    \label{tab:cross-category}
    \vspace{-0.5cm}
\end{table}

\keypoint{Cross-dataset Generalisation:} We further use model trained on QuickDraw dataset to extract feature over TU-Berlin dataset, followed by linear evaluation. Compared within dataset training accuracy of $70.6\%$ ($55.6\%$), we obtain a cross-dataset accuracy (Table \ref{tab:cross-dataset}) of $58.9\%$ ($36.9\%$) on TU-Berlin sketch-images (sketch-vectors) without much significant drop in accuracy, thus signifying the potential of our self-supervised method for sketch data.   
\vspace{-0.2cm}
\setlength{\tabcolsep}{10pt}
\begin{table}[!h]
    \centering
    \footnotesize
    \caption{Cross-dataset (QuickDraw $\mapsto$ Tu-Berlin) recognition accuracy: Model pre-trained on QuickDraw is used to extract fixed latent feature on TU-Berlin, followed by linear model evaluation.}
    \begin{tabular}{c|cc|cc}
        \hline \hline
         & \multicolumn{2}{c|}{Image Space} & \multicolumn{2}{c}{Vector Space} \\
        \cline{2-5}
         & Top-1 & Top-5 & Top-1 & Top-5 \\
        \hline
        MoCo \cite{he2020momentum} & 47.5\% & 62.1\% & -- & -- \\
        SimCLR \cite{chen2020simple} & 47.2\% & 62.0\% & -- & -- \\
        CPC \cite{oord2018representation} & 41.4\% & 60.8\% & 27.7\% & 50.9\% \\
        Ours & 58.9\% & 80.5\% & 36.9\% & 61.7\% \\
        \hline
    \end{tabular}
    \vspace{-0.3cm}
    \label{tab:cross-dataset}
\end{table}

\keypoint{Cross-Task Generalisation:} Both sketch and handwriting are hand-drawn data having similarity in terms of how they are recorded, and represented in image and vector space. Thus, we explore whether a model trained on handwritten data using self-supervised task can generalise over sketches, and vice versa. The pooling stride is adjusted so that, using sketch convolutional encoder, we can get sequential feature, and handwriting convolutional encoder can give feature vector representation on sketch images using global pooling. Following this protocol, we obtain (Table \ref{tab:cross-task}) a reasonable cross-task top-1 accuracy of $37.6\%$ and $33.7\%$ on sketch images and vectors on QuickDraw. Conversely, we get no-lexicon WRA of $28.4\%$ and $26.3\%$ on handwritten offline word images and online word vectors, respectively. 

\vspace{-0.1cm}
\setlength{\tabcolsep}{2pt}
\begin{table}[!h]
    \centering
    \footnotesize
    \caption{Cross-task (Sketch $\leftrightarrow$ Handwriting) generalisation results on extracted fixed latent feature . Lexicon: (L), No-Lexicon: (NL) }
    \begin{tabular}{c|cc|cc|cc|cc}
        \hline \hline
         & \multicolumn{4}{c|}{Sketch (QuickDraw)} & \multicolumn{4}{c}{Handwriting (IAM)} \\
        \cline{2-9}
         & \multicolumn{2}{c|}{Image} & \multicolumn{2}{c|}{Vector} & \multicolumn{2}{c|}{Image} & \multicolumn{2}{c}{Vector} \\
         & Top-1 & Top-5 & Top-1 & Top-5 & L & NL & L & NL \\
        \hline
        Random & 14.6\% & 25.7\% & 11.8\% & 22.9\% & 9.8\% & 6.1\% & 7.1\% & 4.5\% \\
        \hdashline
        CPC \cite{oord2018representation} & 19.7\% & 37.8\% & 17.6\% & 36.9\% & 19.5\% & 12.5\% & 15.7\% & 9.7\% \\
        Ours & 37.6\% & 58.4\% & 33.7\% & 55.8\% & 33.8\% & 28.4\% & 31.6\% & 26.3\% \\
        \hline
    \end{tabular}
    \label{tab:cross-task}
    \vspace{-0.3cm}
\end{table}

\keypoint{Further Analysis:} (i) We have also compared with other backbone CNN network, e.g. AlexNet, where we obtain Top-1 accuracy of $63.3\%$ compared to $71.9\%$ on using ResNet50. This confirms suitability of our design across different backbone architecture. (ii) In our implementation, we perform only basic horizontal flipping and random cropping for augmentation. We also experiment with multiple augmentation strategies mentioned in \cite{chen2020simple}, but notice no significant changes. (iii) Furthermore, following the recent works \cite{xu2018sketchmate, collomosse2019livesketch} that jointly exploits raster sketch-image and temporal sketch-vector for sketch representation, we simply concatenate extracted latent feature of sketch-image and sketch-vector, and evaluate through linear classifier. This joint feature improves the top-1 accuracy to $72.8\%$ compared to $71.9\%$ which uses raster image only. 

\vspace{-0.2cm}
\section{Conclusion}
\vspace{-0.1cm}
We have introduced a self-supervision method based on cross-modal  rasterization/vectorization that is effective in representation learning for sketch and handwritten data. Uniquely our setup provides powerful representations for both vector and raster format inputs downstream. Results on sketch recognition, sketch retrieval, and handwriting recognition show that our pre-trained representation approaches the performance of supervised deep learning in the full data regime, and surpasses it in the low data regime. Thus \modelName{} provides a powerful tool to scale and accelerate deep-learning-based freehand writing analysis going forward.

{\small
\bibliographystyle{ieee_fullname}
\bibliography{Original_egbib}
}

\end{document}